\documentclass[10pt, conference, compsocconf]{IEEEtran}
\usepackage{graphicx}
\usepackage{amsmath}

\ifCLASSINFOpdf
\else
\fi

\begin{document}
%
\title{A Multi-oriented Chinese Keyword Spotter Guided by Text Line Detection}


\author{\IEEEauthorblockN{Pei Xu, Shan Huang, Hongzhen Wang, Hao Song, Shen Huang, Qi Ju}
\IEEEauthorblockA{Tencent Research, Beijing, China 518000\\
\{palinxu, lattehuang, ernestwang, ashersong, springhuang, damonju\}@tencent.com
}

}


%


\maketitle

\begin{abstract}
Chinese keyword spotting is a challenging task as there is no visual blank for Chinese words. Different from English words which are split naturally by visual blanks, Chinese words are generally split only by semantic information.  In this paper, we propose a new Chinese keyword spotter for natural images, which is inspired by Mask R-CNN. We propose to predict the keyword masks guided by text line detection. Firstly, proposals of text lines are generated by Faster R-CNN; Then, text line masks and keyword masks are predicted by segmentation in the proposals. In this way, the text lines and keywords are predicted in parallel. We create two Chinese keyword datasets based on RCTW-17 and ICPR MTWI2018 to verify the effectiveness of our method.
\end{abstract}

\begin{IEEEkeywords}
Chinese keyword; keyword spotting; text line detection;

\end{IEEEkeywords}

%
\IEEEpeerreviewmaketitle

\section{Introduction}
The automatic recognition of text in the images has been a popular topic as text in the images can contain rich information. Word spotting or keyword spotting is aimed to extract specific words from the images, which can be used for geolocation, product search, adding image tags and image filtering etc.

\begin{figure}[ht]
\centering
\includegraphics[width=0.9\linewidth]{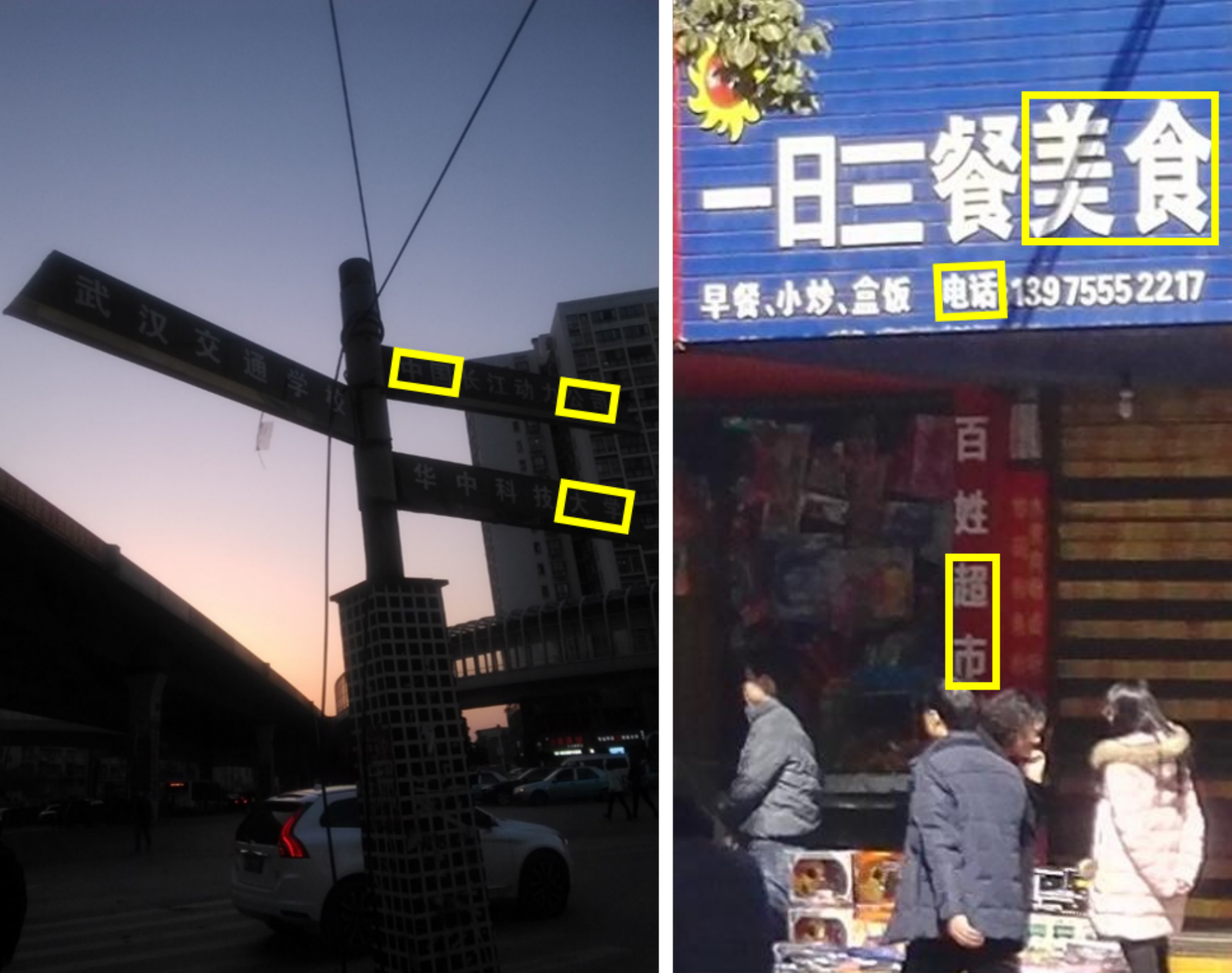}
\vspace{-2mm}
\caption{Illustration of Chinese keyword spotting. We need to find all keywords in the images with their categories and localization. As shown, the text can be in various orientations. The yellow boxes in the images are the Chinese keywords, which mean ``China'',``Company'', ``University'', ``Food'', ``Telephone'', and ``Supermarket'' respectively.}
\label{fig:introduction}
\end{figure}

Most of the previous works~\cite{DBLP:journals/pr/AlmazanGFV14,DBLP:journals/pami/AlmazanGFV14,DBLP:conf/icdar/RothackerF15,DBLP:conf/iccv/WilkinsonLB17} focused on English keyword spotting. In this paper, we focus on Chinese keyword spotting. As shown in Fig.~\ref{fig:introduction}, there are some challenges in Chinese keyword spotting.
1) The keywords or text lines can be in various orientations or shapes, including horizontal, oriented, and vertical.
2) Different from English words naturally split by visual blanks, Chinese words are generally split only by semantic information because there are no visual blanks. Thus, it is hard to directly obtain the word bounding boxes directly.
3) The annotations of Chinese characters or keywords are expensive. For the characters, it is hard to annotate all of them due to the large amounts; For the keywords, they can be differently defined in different tasks.

 Inspired by Mask TextSpotter~\cite{lyu2018mask} which combines the word detection and the character segmentation into a unified network, we propose to segment the keyword guided by the text line proposals. We first obtain the proposals of text lines from Faster R-CNN. Then the text line mask branch and keyword mask branch share the proposals to segment text lines and keywords parallelly. The whole architecture of our model is inherited from Mask R-CNN~\cite{he2017mask} and can be trained in an end-to-end manner. Due to the accurate prediction of segmentation and synthesized text images in arbitrary orientations, our method is able to handle the keyword of various shapes, including horizontal, oriented, and vertical shapes. Moreover, the proposed model only needs text line level annotations of real-world images since we can synthesize text images and train in a semi-supervised way.
 
The contributions of this paper can be summarized as follows:
1) We proposed an end-to-end trainable keyword spotter which can handle keywords of different orientations.
The pipeline is simple and efficient since the keyword and the text line can be segmented from text line proposals in parallel without recognition.
2) Our proposed method can spot Chinese keywords which have no visual blanks.
3) We only need the annotations of text lines in the real-world images. The keyword annotations can be obtained easily by synthesizing corresponding text images.  


\section{Related Work}
\subsection{Keyword Spotting}
Previous Quey-by-String (QbS) keyword spotting works can be roughly classified into two categories: segmentation-based methods and segmentation-free methods. 

Segmentation-based methods~\cite{DBLP:journals/pami/AlmazanGFV14,DBLP:journals/pami/FrinkenFMB12,DBLP:conf/icdar/ZhangZL13,zhang2014character,DBLP:journals/eaai/ZhangZL17} firstly segment the all the words and then transform the word feature vectors and the queries into the same space by word embedding. Segmentation-based methods are able to deal with a large set of keywords, but limited to the effectiveness of the segmentation and needs expensive annotations in word, character or radical level, which is an especially severe problem for Chinese keyword spotting. Especially in Chinese text, word segmentation is a challenging task as there are no visual blanks between the words. The words in Chinese are generally segmented by semantic meanings. Besides, methods like [8][9][10] mainly focus on handwritten Chinese documents and can only handle horizontal text, which are hard to deal with arbitrary orientation and composition in scene images. Thus, previous segmentation-based methods are not suitable for Chinese keyword spotting in scene images.

Segmentation-free methods~\cite{DBLP:journals/pr/AlmazanGFV14,DBLP:conf/icdar/RothackerF15,DBLP:conf/icdar/GhoshV15,DBLP:conf/iccv/WilkinsonLB17} usually generate a lot of proposals in the whole image. Then, these proposals are matched with the query strings. For example, \cite{DBLP:journals/pr/AlmazanGFV14} is a segmentation-free method based on Bag of Word (BoW) and SIFT descriptors~\cite{lowe2004distinctive}. Segmentation-free methods do not need to segment all the words and have the set of keywords are flexible. The main drawback with these methods is the large amounts of regions generated, resulting in false positives and low efficiency.

Our method does not segment all the words thus it is more like a segmentation-free method. It share proposals for text lines and keywords, and only spots keywords from the text proposals. In this way, it does not need to search keywords in the whole image, which is a benefit to efficiency. Meanwhile, it is suitable for Chinese text where there are no visual blanks between the neighbor words.

\subsection{Differences From Scene Text Spotting}

There are a lot of works on scene text spotting showing effectiveness in English text. Jaderberg et al.~\cite{DBLP:journals/ijcv/JaderbergSVZ16} propose a practical system using CNNs in a very comprehensive way. Liao et al.~\cite{liao2017textboxes,liao2018textboxes++} use a single-shot text detector along with a text recognizer to detect and recognize horizontal and oriented scene text in images respectively. ~\cite{li2017towards,busta2017deep} start to integrate the scene text detection and recognition modules into a unified framework. \cite{lyu2018mask} propose an end-to-end trainable scene text spotter enjoying a simple pipeline and can handle text of various shapes. 

However, these methods have not shown their effectiveness in Chinese scene text spotting yet, which is still a problem to be solved. Besides, scene text spotting is required to detect and recognize almost all the words in the image, making the system complex and the pipeline long. However, keyword spotting is quite different from scene text spotting. We only care about the target keywords in images, which is similar to voice wake-up, enabling us to exploit a simpler system and shorter pipeline for efficiency.

Our method is inspired by some of the scene text spotting methods. It has the following strengths against OCR methods: 1) it has a simpler and shorter pipeline, which predicts text lines and keywords in parallel, no needing to recognize the text line. 2) it can give precise location of keywords while most of the existing sequence-to-sequence OCR methods have difficulty in alignment.

\section{Methodology}
\subsection{Overview }

\begin{figure*}[ht]
\centering
\includegraphics[width=0.9\linewidth]{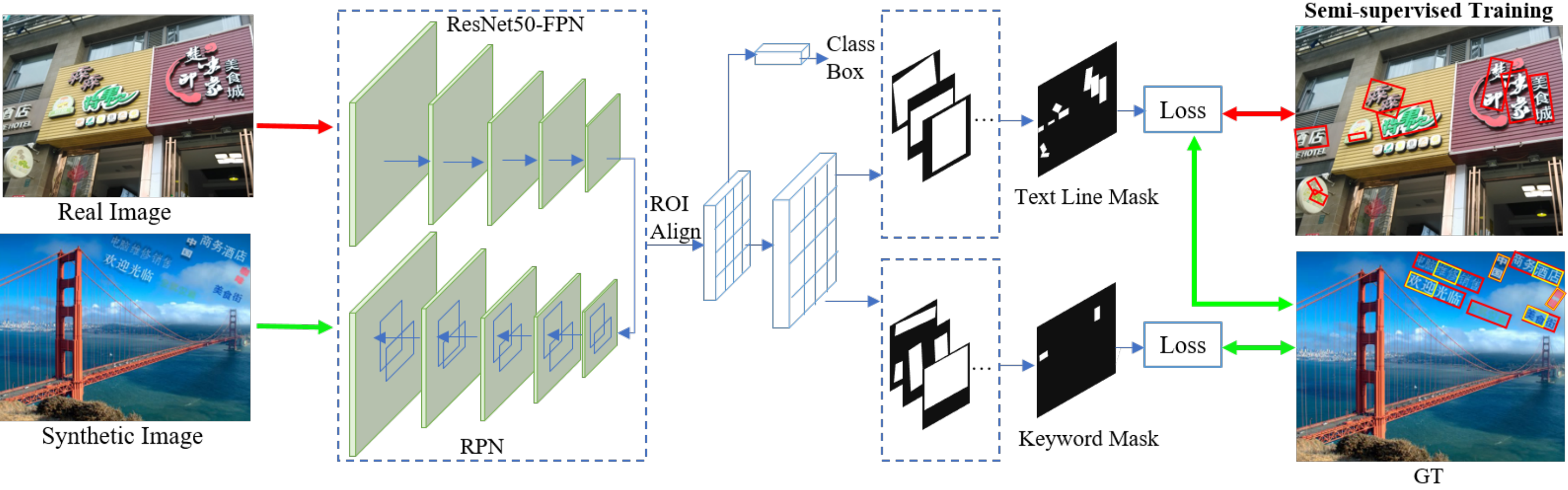}
\vspace{-4mm}
\caption{Architecture of our proposed keyword spotter. It consists of a Faster-R-CNN framework and two branches for text line segmentation and keyword segmentation. The boxes in red and yellow mean the ground truth of text lines and keywords respectively. The arrows in red and green mean the training for real images and synthetic images respectively. We only use text line annotations of real images in a semi-supervised way.}
\label{fig:architecture}
\end{figure*}

The overall architecture of our proposed method is illustrated in Figure~\ref{fig:architecture}, which is based on the Mask-RCNN~\cite{he2017mask}. It mainly consists of a Faster R-CNN~\cite{ren2015faster} framework for text lines detection and two instance segmentation branches that predict the text line mask and keyword mask in parallel. Specifically, in the Faster R-CNN framework, we adopt a ResNet-50 FPN network as our backbone and a region proposal network (RPN) to generate text line proposals, from which the ROI features are extracted with ROI align to refine the proposals by box classification and regression. The sharing ROI features are fed into the text line instance segmentation branch and the keyword instance segmentation branch to get text line masks and keyword masks in parallel. Rotated rectangle boxes of text line and keyword can be obtained by a simple post-processing algorithm from the corresponding masks.

\subsection{Text Line Region Proposal Network }

Different from most common word-level detection and spotting in English text images, such as ICDAR 2013, ICDAR 2015, etc., Chinese text can hardly be detected in word level, since there are no visual blanks. Chinese text lines naturally have variety in length as there is no limit to the number of characters in a text line. Besides, Chinese text lines have a large variety of sizes and orientation. Thus, we specially design the aspect ratios and sizes of the anchors to solve this problem. Specifically, we set the aspect ratios of RPN to (0.1, 0.2, 0.5, 1, 2, 5, 10) in each stage to better match text lines which are long. Following Mask R-CNN~\cite{he2017mask}, we adopt an FPN backbone to extract multi-level RoI features from different levels of the feature pyramid according to anchor scale (32, 64, 128, 256, 512) in five stages of RPN network respectively. All the proposals are sampled to $7 \times 7$  features by RoI align for box classification and regression and $14 \times 14$ features for subsequent instance segmentation.

\subsection{Multi-Task Branch for Instance Segmentation}
\subsubsection{Text Line Mask Branch}
 
The text line mask branch is used to predict accurate localization of text lines.
In the text line branch, we keep the aspect ratio of RoI features for segmentation as $1$ since the text line may be horizontal or vertical. Specially, The extracted RoI features with the size $14 \times 14$ are fed into four convolutional layers and a de-convolutional layer upsampling to instance segmentation masks in the size of $28 \times 28$. The masks include 2 maps for background and text line instance these two classes respectively.

\subsubsection{Keyword Mask Branch}

In our task, keywords are in the text lines, where characters are adjacent with uncertain char spacing but without visual blanks. 
Thanks to line-level spotting methods, we can simply add a keyword mask branch after the RoI features in parallel with the text line mask branch, thus obtaining the keyword regions only from text regions. Besides, this branch brings negligible extra computation and time consumption. This branch has four convolution layers and de-convolutional layers similar to text line branch, outputting K+1 keyword instance segmentation masks in the size of $28 \times 28$, where 1-K is for keyword maps and 0 for background map. In our experiments, the K is 30, i.e., we selected 30 keywords for this task. Note that segmentation can often suffer adhesion problem when two keywords are adjacent, especially in small keywords. So we adopt a shrink strategy on generating keyword label to solve this problem. In our experience, we set the shrinking scale as 0.8 since Chinese keywords are usually 2-4 characters. The loss of keyword mask branch for a proposal is as follow:
\begin{equation}
\label{eq:loss}
L_{key} = -\frac{1}{N}\sum_{n=1}^{N}\sum_{k=1}^{K} Y_{n,k} log(\frac{e^{X_{n,k}}}{\sum_{j=0}^{K-1} e^{X_{n,j}}}),
\end{equation}
where $K$ is the number of classes, $N$ is the number of pixels in each map and $Y$ is the corresponding ground truth of the output maps $X$.

\subsection{Image Synthesis}
Almost all of keyword spotting tasks use word annotations in English handwriting document images or scene text images, and char annotation in Chinese text images. However, the annotations cost lots of efforts especially when text is dense on images. So we adopt a keyword image synthesis engine based on the Gupta image synthesis~\cite{gupta2016synthetic} to create our training datasets. Firstly, we add 4000 non-text image datasets to the background images. Secondly, we improve the synthesis engine to generate vertical Chinese text lines, which is rarely seen in English text lines. And we can enlarge the probability to generate text lines with different extreme orientations. Finally, we refine the text rendering as follow: 1) adding multiple Chinese fonts and establish a char-to-font dictionary, as some chars cannot render successfully in some fonts. 2) adjusting the font color according to text color clusters of real images. 3) some modifications in opacity, blur and depth model. 

\begin{figure*}[ht]
\centering
\includegraphics[width=0.9\linewidth]{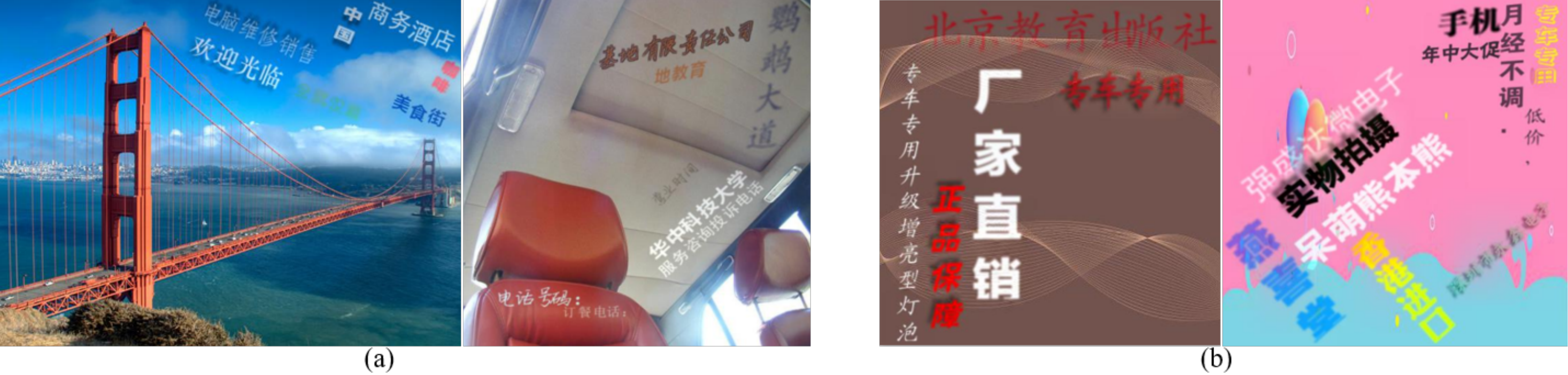}
\vspace{-3mm}
\caption{Synthetic images. (a) synthetic images for RCTW17, mainly text images in the wild; (b) synthetic images for MTWI2018, mainly digit-born images.}
\label{fig:synthesis}
\end{figure*}

\subsection{Semi-supervised Training Guided by Text Line Detection}
In the training process, we first use synthetic images to pretrain the model and then finetune on the real dataset. However, in many circumstances, it is hard to get keyword annotations in real images without char-level annotations. Inspired by Mask TextSpotter~\cite{lyu2018mask}, we introduce a semi-supervised training scheme only guided by text line annotation of real images in the finetuning procedure. Specifically, we put synthetic images and real images together with a ratio to finetune. For synthetic images, we update the losses of text line mask branch and keyword mask branch normally. For real images without keyword annotations, we only update the loss of text line mask branch and set the loss of keyword mask branch to 0 as follow: 
\begin{equation}
  L_{key'} = 
  \begin{cases}
    L_{key}& \text{if it is a synthetic image}, \\
    0& \text{otherwise}
  \end{cases}
\end{equation}
where $L_{key}$ is illustated in Eq.~\ref{eq:loss}.

Note that all the images are mixed up in a random order. By the method of semi-supervised training, we can achieve higher keyword spotting performance on the real dataset with the text line detection performance increasing. 

\section{Experiments}
\subsection{Chinese Keyword Datasets Creation}
keyword spotting is a task related to text localization and recognition. Since there are some popular Chinese scene text reading datasets but almost none Chinese keyword dataset, we create keyword datasets from them. However, they only provide line-level ground-truths for the training set and no ground-truths for testing set. So We can only create our keyword dataset based on these training set.

\textbf{RCTW17-Keyword}: RCTW-17 is a Chinese scene text reading dataset containing 8034 training images taken from wild streets and indoor scenes, involving lots of complex in text detection and recognition. The images have a great size varying from 500x500 to 3000x4000 and are annotated inline-level without char or word annotations. We extract 30 meaningful keywords with high frequency from all the annotated text， which are representative of landmarks, geolocation, signs, and neighborhoods. There are 2781 images containing these 30 keywords, which becomes our test set.

\textbf{MTWI2018-Keyword}: ICPR MTWI2018 is a challenging dataset containing 10000 training images from the web, including digital-born image, product description and web ad image, most of which involve complex typography, dense small text or watermarks. The images vary from 500x500 to 1000x1000 in size and are annotated with text line. We also extract 30 meaningful keywords with high frequency from all the annotated text, which are representative of advertisement slogans and product information. There are 5306 images containing these 30 keywords as a test set.

\subsection{Evaluation Protocols}
We use the mean average precision (mAP) as the accuracy measure, which is the standard measure of performance for retrieval tasks on overall query classes. We can calculate average precision (AP) on every keyword class and get mAP by averaging. Note that the AP is acquired with different score threshold to the detected keywords of each image. The measurement is as follow:
\begin{align}
AP &= \frac{1}{|r|}\sum_{j=1}^{N}p(j)r(j)\\
mAP &= \frac{\sum_{q=1}^{K}AP(q)}{|K|}
\end{align}
Where $p(j)$ is the precision measured at cut-off $j$ in the returned list and $r(j)$ is 1 if a return list at rank $j$ is relevant, otherwise 0. In practice, for retrieved images of each keyword class sorted by confidence, an image is relevant if its ground-truth contains the keyword else 0.

\subsection{Model Comparisons}
\subsubsection{baseline}
The baseline is only trained with synthetic images, without utilizing annotations of text lines in real images.
\subsubsection{baseline+guided}
The ``baseline+guide'' is first trained with synthetic images and then finetuned with a mixture of the synthetic images and the real images with text-line level annotations in a semi-supervised way.
\subsubsection{TextBoxes++}
This is a comparison experiment based on TextBoxes++, which detects keyword candidates in the whole images. It cannot utilize text line detection of real images since it only aims at predicting keyword boxes but not text lines.

\begin{figure*}[ht]
\centering
\includegraphics[width=0.95\linewidth]{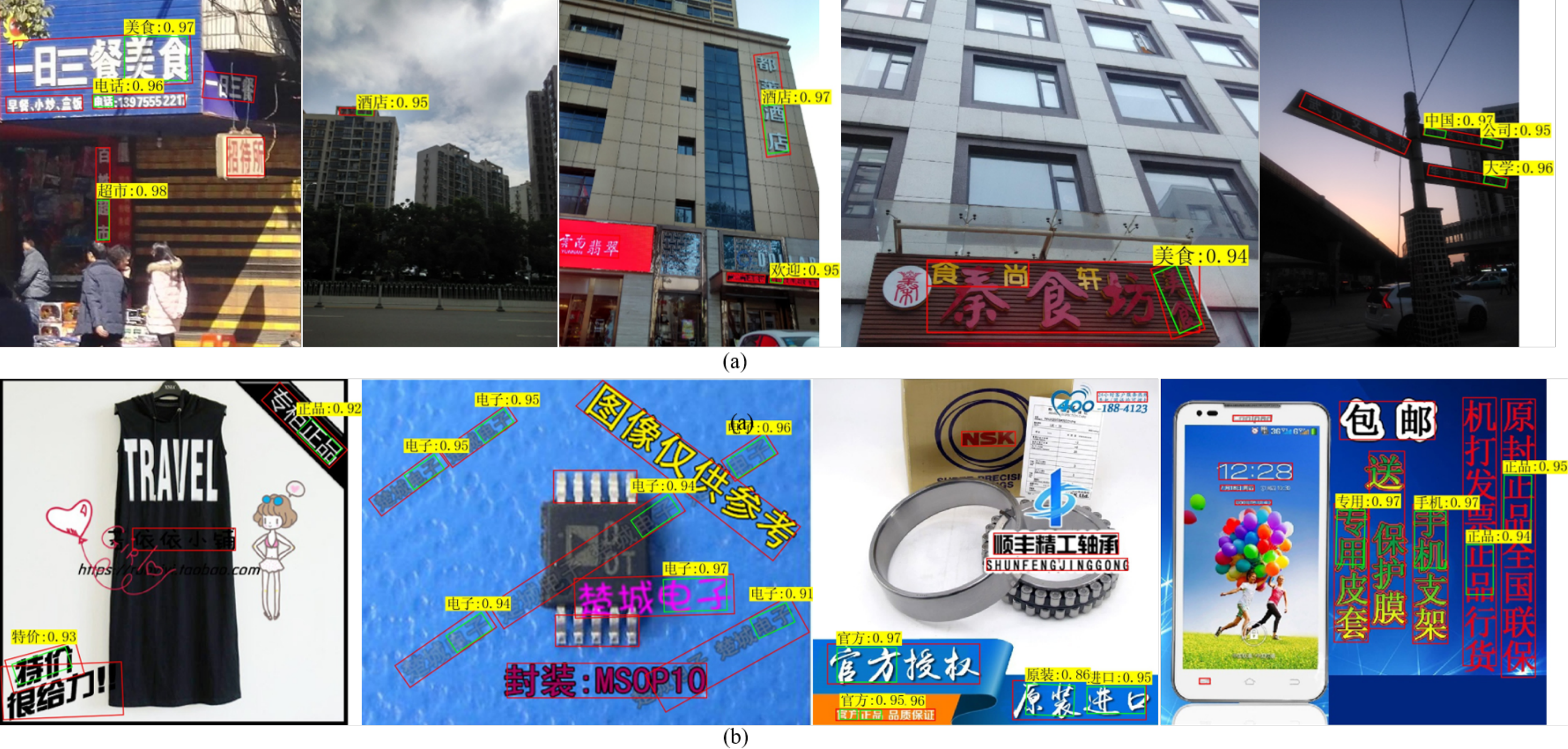}
\vspace{-3mm}
\caption{Visualization of keyword spotting results. (a) visualization of results on RCTW17-Keyword; (b)visualization of results on MTWI2018-Keyword. Detected text lines are in red and spotted keywords are in green with confidence.}
\label{fig:results}
\end{figure*}

\subsection{Experimental Results}
\textbf{RCTW-17 keyword dataset}: Results are shown in Tab.~\ref{table:RCTW17}. When only training on the synthetic images, our method can achieve mAP $0.7049$, which is nearly $3$ points higher than $0.6782$ mAP with TextBoxes++. It is reasonable that our method can utilize implicit relationships between text lines and keywords, thus contributing to reduce false positives in the whole image. Furthermore, when we finetune with real images mixing with synthetic in a semi-supervised way, we can see an obvious gain in keyword spotting performance, achieving $0.7955$. Different mixing ratios between synthetic images and real images also make a difference, where mixing ratio 2:1 gives the best results. 

\begin{table}[ht]
\footnotesize
\centering
\begin{tabular}{|c|c|c|}
\hline
Method                        & mixing ratio & mAP   \\ \hline
TextBoxes++ ($768\times768$)                & - & 0.6572        \\ \hline
TextBoxes++ ($1024\times1024$)                & - & 0.6782         \\ \hline
Baseline                      & - & 0.7049         \\ \hline
Baseline+guided                 & 16:1 & 0.7875         \\ \hline
Baseline+guided           & 8:1 & 0.7949      \\ \hline
Baseline+guided           & 2:1 & \textbf{0.7955}    \\ \hline
\end{tabular}
\caption{Evaluation results of several methods with different settings on RCTW17-Keyword dataset. 
\label{table:RCTW17}
}
\end{table}

\textbf{MTWI2018 keyword dataset}: Results are shown in Tab.~\ref{table:MTWI2018}. Our method can achieve mAP $0.8937$ on MTWI2018-Keyword dataset when training on the synthetic data, higher than $0.8826$ with TextBoxes++. The advantage becomes not obvious for the feature of MTWI2018-Keyword dataset. Specially, the challenges of MTWI2018-Keyword lie in the complex typography and dense small text, nevertheless the background is simple and the font is clear. Thus, text line detection of MTWI2018 is more affected than RCTW17 thus drawing back the keyword spotting performance. After we finetune with real images mixing with synthetic in a semi-supervised way, our keyword spotting performance can achieve $0.9212$. Mixing ratio as 2:1 give the best results. 

Figure~\ref{fig:results} shows visualizations of some results on RCTW-17 Keyword and MTWI2018-Keyword respectively.

\begin{table}[ht]
\footnotesize
\centering
\begin{tabular}{|c|c|c|}
\hline
Method                        & mixing ratio & mAP   \\ \hline
TextBoxes++                & - & 0.8826         \\ \hline
Baseline                      & - & 0.8937         \\ \hline
Baseline+guided                 & 16:1 & 0.9127         \\ \hline
Baseline+guided                 & 8:1 & 0.9158         \\ \hline
Baseline+guided           & 2:1 & \textbf{0.9212}       \\ \hline
\end{tabular}
\caption{Evaluation results of several methods with different settings on MTWI2018-Keyword dataset. 
\label{table:MTWI2018}
}
\end{table}

\subsection{Implementation Details}
We synthesize 130,000 images for RCTW17-Keyword and 160,000 images for MTWI2018-Keyword with text line and char annotations, which become our training datasets. Some synthetic image examples are shown in Figure~\ref{fig:synthesis}.

For experiments on RCTW17-Keyword, we train TextBoxes++ on synthetic images for $38k$ iterations with batch size $32$, learning rate $0.001$. The learning rate is decreased to tenth after $30k$ iterations. The training scales is $384 \times 384$ and testing scales are $768 \times 768$. We train our baseline on synthetic images for $32.5k$ iterations with batch size $16$, learning rate $0.01$. The shorter side of the training images are set to 800, keep the aspect ratio of the images. In the inference period, the shorter side of the input images are set to 1200. our baseline+guided is finetuned for $1.5k$ iterations with mixing ratio 2:1 between synthetic images and real RCTW17 images, learning rate 0.0005. 

For experiments on MTWI2018-Keyword, we train TextBoxes++ on synthetic images for $33k$ iterations with batch size $32$, learning rate $0.001$. The learning rate is decreased to tenth after $20k$ iterations. The training and testing scales are $768 \times 768$. We train our baseline on synthetic images for $37.5k$ iterations with batch size $16$, learning rate $0.01$ which is decreased to tenth at the $35k$ iteration. The scale of the input images are the same as these in RCTW17-Keyword. Our baseline+guided is finetuned for $1.5k$ iterations with mixing ratio 2:1 between synthetic images and real MTWI2018 images, learning rate 0.0005. The training and testing scales are both 800 in the shorter side.

\begin{figure}[ht]
\centering
\includegraphics[width=0.9\linewidth]{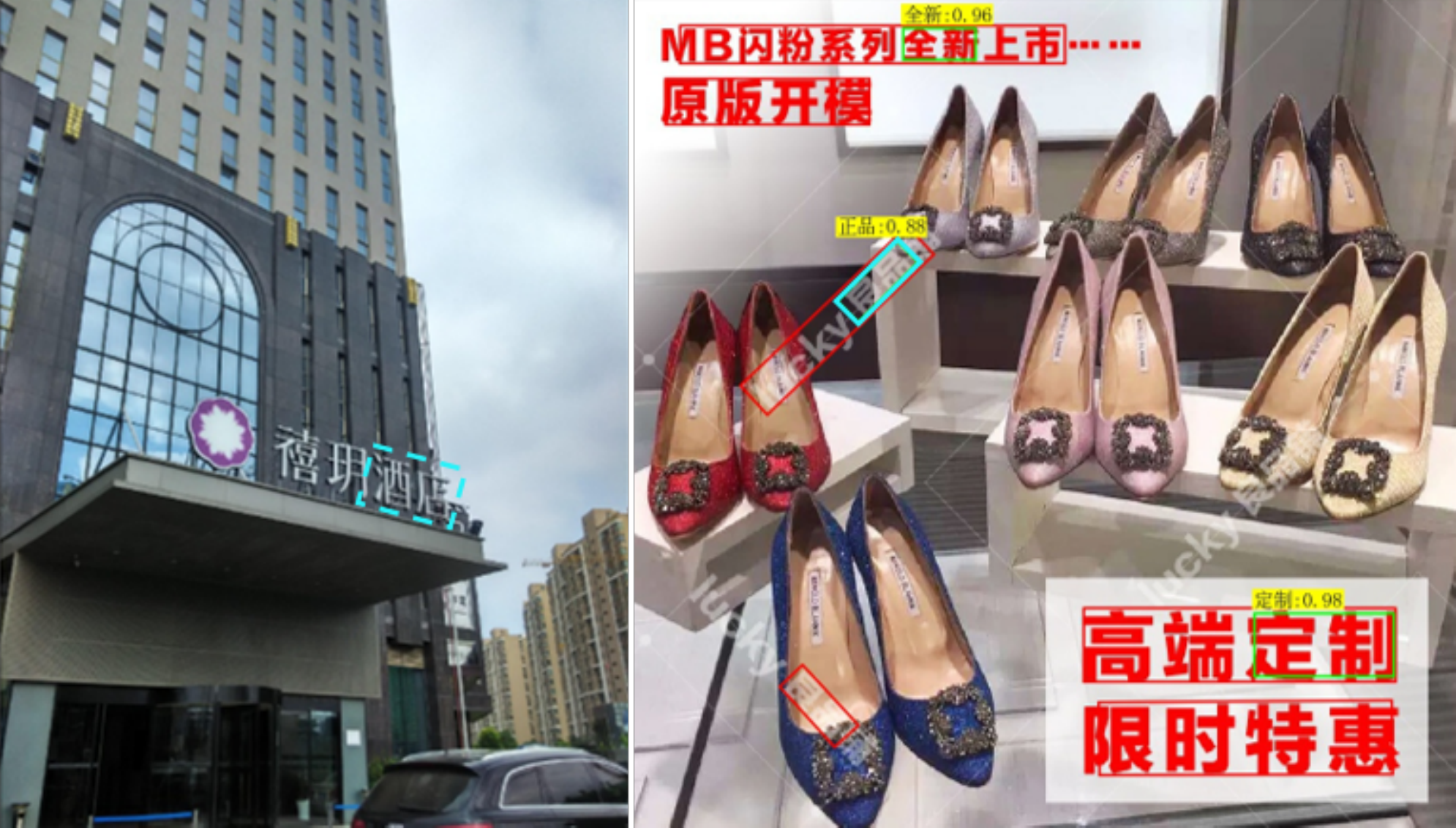}
\vspace{-2mm}
\caption{Failure cases. The missing boxes are in dash cyan and the false positive boxes are in solid cyan.}
\label{fig:failures}
\vspace{-5mm}
\end{figure}

\subsection{Limitation}
There are some failure cases of our method, as we show in Figure~\ref{fig:failures}. Our method detects wrong words which have similar radicals and structures to a keyword, in the fact that only about 500 radicals are adequate to describe more than 20,000 Chinese characters. Another failure case is when the text line is hard to detect and then the keyword may be missed. More real and complex synthetic images and fine-grained information may be required to address this issue.

\section{Conclusion}
In this paper, we proposed a Chinese keyword spotter which is guided by text line detection. The proposed method is able to handle the keywords of various shapes, including horizontal, oriented, and vertical shapes. Benefited from the synthetic data and the good generality of the proposed method, only text line level annotations are needed for real-world training images. We proved the effectiveness of our method by conducting experiments on two keyword datasets which are created from two scene text benchmarks.





\bibliographystyle{IEEEtran}
\bibliography{reference}
%



\end{document}